\documentclass{easychair}

\usepackage{booktabs} 
\usepackage{subcaption}
\usepackage[utf8]{inputenc} % allow utf-8 input
\usepackage[T1]{fontenc}    % use 8-bit T1 fonts
\usepackage{hyperref}
\usepackage{url}            % simple URL typesetting
\usepackage{booktabs}       % professional-quality tables
\usepackage{amsfonts}       % blackboard math symbols
\usepackage{nicefrac}       % compact symbols for 1/2, etc.

\usepackage{graphicx} % margin definition
\usepackage{pgfplots} % graphics
\pgfplotscreateplotcyclelist{rw}
{solid, mark repeat = 80, mark phase = 80, mark = *, black\\
 solid, mark repeat = 80, mark phase = 80, mark = square*, black\\}

\usepackage{tikz} % graphics
\usetikzlibrary{shapes,arrows}
\usepackage{pgfplots}
\usepackage{float}
\usepackage{bussproofs}
\usepackage{xspace}
\usepackage{amsmath}
\usepackage{amssymb}
\usepackage{mathtools}
\usepackage{doi}
\usepackage{multicol}

\def\holfour{\textsf{HOL4}\xspace}

\def\vampire{\textsf{Vampire}\xspace}
\def\eprover{\textsf{E-prover}\xspace}

\def\tactictoe{\textsf{TacticToe}\xspace}
\usepackage{color}

\usetikzlibrary{arrows,shapes,calc}
\usetikzlibrary{trees,positioning,fit}
\tikzstyle{arrow}=[draw,-to,thick]
\tikzstyle{embedding} = [draw, minimum width=15mm, minimum height=6mm]
\tikzstyle{nnop} = [draw, minimum width=8mm, minimum height=8mm, rounded 
corners]
\tikzstyle{block} =
  [rectangle, draw,
   minimum width=9mm, text centered, rounded corners, minimum height=2.0em]
\tikzstyle{line}=[draw]
\tikzstyle{cloud} =
  [draw, text centered, ellipse, minimum height=2.0em, minimum width=9em]
\usepackage[]{listings}
\lstdefinelanguage{SML}
{columns=fixed,
  keywords={THEN,THENL,val,open,store_thm,fun,fn,let,in,end,true,
  while,do,if,then,else,break,return,;},%
  frame=none,
  sensitive=true,
  keywordstyle=\fontfamily{lmss}\scriptsize\selectfont,%
  basicstyle=\fontfamily{pcr}\small\selectfont,%
  stringstyle=\tt,
  morestring=[b]",
  literate=
   {=}{{\tt\raisebox{-.15mm}{=}}}1%
   {[}{{\tt\raisebox{-.15mm}{[}}}1%
   {]}{{\tt\raisebox{-.15mm}{]}}}1%
   {->}{{$\rightarrow$}}1%
   {wedge}{{$\wedge$}}1%
   {vee}{{$\vee$}}1%
   {==>}{{$\Rightarrow$}}1%
   {<==>}{{$\Leftrightarrow$}}1%
   {vee}{{$\vee$}}1%
   {<=}{{$\leq$}}1%
   {>=}{{$\geq$}}1%
   {'a}{{$\alpha$}}1%
   {'b}{{$\beta$}}1%
   {ldots}{$\ldots$}1%
   {emptyset}{$\emptyset$}1%
   {!}{$\forall$}1%
   {fff}{{DB.fetch}}1%
   {``}{\hspace{-.5mm}`\hspace{-1mm}`}1%
  }

\title{Deep Reinforcement Learning for\\ Synthesizing Functions in Higher-Order 
Logic
\thanks{This work has been supported 
by the European Research Council (ERC) grant
AI4REASON no. 649043  under the EU-H2020 programme. 
We would like thank Josef Urban for his contributions to the first version of 
this paper.}}

\author{Thibault Gauthier}
\institute{Czech Technical University in Prague, Prague, Czech Republic\\
\email{email@thibaultgauthier.fr}}
\authorrunning{T. Gauthier}
\titlerunning{Deep Reinforcement Learning for Synthesizing Functions in 
Higher-Order 
Logic}

\begin{document}
\maketitle
\begin{abstract}
The paper describes a deep reinforcement learning framework based on
self-supervised learning within the proof assistant HOL4. A close 
interaction between the 
machine learning modules and the HOL4 library is achieved by the choice of tree 
neural networks (TNNs) as machine learning models and the internal use of 
HOL4 terms to represent tree structures of TNNs. Recursive improvement is 
possible when a task is expressed as a search problem.
In this case, a Monte Carlo Tree Search (MCTS) algorithm guided by a TNN can be 
used to explore the search space and produce better examples for training the 
next TNN. As an illustration, term synthesis tasks on combinators and 
Diophantine equations are specified and learned. 
We achieve a success rate of 65\% on combinator synthesis problems 
outperforming state-of-the-art ATPs run with their best general set of 
strategies. We set a precedent for statistically guided synthesis of 
Diophantine equations by solving 78.5\% of the generated test problems.
\end{abstract}

\section{Introduction}
Improvements in automated theorem provers (ATPs) have 
been so far predominantly done by 
inventing new search paradigms such as 
superposition~\cite{DBLP:journals/tcs/Fribourg85} and 
SMT~\cite{DBLP:series/faia/BarrettSST09}. Over the 
years, developers of these provers have optimized their modules and fine-tuned 
their parameters. As time progresses, it is becoming evident that more 
intricate collaboration between search algorithms and intuitive guidance is 
necessary.
ATP developers have frequently manually translated their intuition into 
guiding heuristics and tested many different parameter combinations. The first 
success demonstrating the possibility 
of replacing these heuristics by machine learning guidance has been 
demonstrated in ITP Hammers~\cite{hammers4qed}. There, feature-based predictors 
on large interactive theorem prover (ITP) libraries learn to select relevant 
theorems for a conjecture. This step drastically reduces the search space.
As a result, ATPs can prove many conjectures proposed by ITP users.
The last landmark that is a major source of inspiration for this paper 
is the development and success of self-improving neurally guided algorithms in 
perfect information games~\cite{silver2017mastering}. In this work, we adapt 
such algorithms to theorem proving tasks. 
 
The hope is that systems using the learning paradigm  will eventually improve 
on and outperform the best 
heuristically-guided ATPs.
We believe that the best design for a solver is one that generalizes
across many tasks. One way to achieve this is to minimize the amount of 
algorithmic bias based on human knowledge of the specific domain.
Eventually, given enough examples, the neural architecture might be able to 
recognize and exploit patterns by itself. For large domains that require vast 
amount of knowledge and understanding, the number of required examples to 
capture all the patterns is too large. That is why our experiments 
are performed on a domain that contains a small number of 
basic concepts. 

\paragraph{Term Synthesis Tasks}
To test our approach, we choose among theorem proving tasks two 
term synthesis tasks. In itself, term synthesis is a less commonly explored 
technique as it is often a less
efficient way of exploring a search space as deduction-based methods. 
This technique is however crucial in inductive theorem proving~\cite{hipspec13} 
and in counterexample 
generators~\cite{DBLP:conf/cpp/Bulwahn12,DBLP:conf/itp/BlanchetteN10} 
as it can be used to provide an induction predicate or a witness.
%The current approaches for such tasks often rely on brute force enumeration, 
%heuristics and manually deduced constraints.

A term synthesis task can be expressed as proving a theorem of the form
$\exists x.\ P(x)$ with the proof providing a witness for the term $x$. 
In both our tasks, the theorem can be re-stated as $\exists x.\ f(x) = y$, 
where $f$ is an evaluation function specific to the task and 
$y$ is an image specifying the particular instance that has to be solved.
In this light, the aim of the prover is to find an element of the preimage of 
$y$.
This might be hard even if we have an efficient algorithm for $f$ as 
conjectured by the existence of one-way 
functions~\cite{DBLP:journals/mlcs/Gradel94}.
%Moreover in the two tasks, the domain of $f$ is a set of functions 
%(combinators 
%or polynomial), thus the evaluation function $f$ is higher-order.

In the first task, the aim is to construct an untyped SK-combinator which 
is semantically equal to a $\lambda$-expression in head-normal form. 
Applications range from better encoding of $\lambda$-expressions in 
higher-order to first-order 
translations~\cite{sledgehammer10,DBLP:conf/cpp/Czajka16}
to efficient compilations of functional programming languages to combinator code
~\cite{DBLP:journals/cl/JoyRB85}.

In the second task, the aim is to construct a polynomial $p$ whose Diophantine 
set $D(p)$ is equal to a set of integers $S$. We say that the witness $p$ 
describes $S$. This process of constructing new and equivalent definitions for 
$S$ is important during mathematician investigation as it gives alternative 
point of view for the object $S$.  For instance, the set of natural number 
$\lbrace 0,2,4,8,10,12,\ldots\rbrace$ can be describe by the Diophantine 
equation $k = 2\times x$ defining the concept of even numbers.
The Encyclopedia of integer sequences~\cite{DBLP:conf/mkm/Sloane07} contains 
entries that can be 
stated as Diophantine equations (e.g. sequence A001652).
Other motivations for investigating Diophantine equations
comes from elliptic curve cryptography~\cite{DBLP:phd/dnb/Baier02} and number 
theory~\cite{duverney2010number}.

\paragraph{Contributions}
This paper presents a general framework that lays out the foundations for 
neurally guided solving of theorem proving tasks. 
We evaluate the suitability of tree neural networks on two term synthesis tasks.
We focus on showing how deep 
reinforcement learning~\cite{Sutton:1998:IRL:551283} algorithms can acquire the 
knowledge necessary to solve 
such problems through exploratory searches. 
The framework is integrated in the \holfour~\cite{hol4} system (see 
Section~\ref{sec:result}). 
%Hence, \holfour terms and procedures are used to specify our tasks.
The contributions of this paper are: 
(i) the implementation of TNNs with the associated backpropagation algorithm,
(ii) the implementation of a guided Monte Carlo Tree Search (MCTS) 
algorithm~\cite{montecarlo}
for arbitrarily specified search problems,
(iii) the demonstration of continuous self-improvement for a large number of 
generations,
(iv) a comparison with state-of-the-art theorem provers on the 
task of synthesizing combinators and
(v) a formal verification of the solutions in \holfour.

\section{Tree Neural Networks}\label{sec:tnn}
In the machine learning field, various kinds of predictors are more 
suitable for learning various tasks. That is why with new problems come new 
kinds of predictors. It is particularly true for predictors such as
neural networks. For maximum learning efficiency, the structure of the problem 
should be reflected in the structure of the neural network. For example, 
convolutional neural networks are best for handling pictures as their structure 
have space invariant properties whereas recurrent networks can handle text 
better. For our purpose, we have chosen neural networks that are in particular 
designed to take into account the tree structure of terms and 
formulas as in~\cite{DBLP:journals/tacl/KiperwasserG16a}.

\subsection{Architecture}
\begin{figure}
\centering
\begin{tikzpicture}[scale=0.8, every node/.style={scale=1.0}]
\node [embedding,node distance=3cm] (0l) {$a$};
\node [node distance=1.5cm,right of=0l] (0lm) {};

\node [embedding, node distance=3cm, right of=0l] (0m) {$b$};
\node [embedding, node distance=3cm, right of=0m] (0r) {$a$};

\node [nnop, node distance=1cm, above of=0lm] (p) {$f$};
\node [nnop, node distance=1cm, above of=0r] (s) {$g$};

\node [embedding, node distance=1cm, above of=s] (se) {$g(a)$};
\node [embedding, node distance=1cm, above of=p] (pe) {$f(a,b)$};
\node [node distance=2.25cm, right of=pe] (pr) {};
\node [nnop, node distance=1cm, above of=pr] (t) {$f$};

\node [embedding, node distance=1cm,above of=t] (te) {$f(f(a,b),g(a))$};
\node [nnop, node distance=1cm,above of=te] (h) {$\mathit{head}$};
\node [node distance=1cm,above of=h] (he) {};

\draw[-to,thick] (0l) to (p);
\draw[-to,thick] (0m) to (p);
\draw[-to,thick] (0r) to (s);
\draw[-to,thick] (0r) to (s);
\draw[-to,thick] (p) to (pe);
\draw[-to,thick] (s) to (se);
\draw[-to,thick] (pe) to (t);
\draw[-to,thick] (se) to (t);
\draw[-to,thick] (t) to (te);
\draw[-to,thick] (te) to (h);
\draw[-to,thick] (h) to (he);
\end{tikzpicture}
\caption{Computation flow during the application of a TNN to the term 
$f(f(a,b),g(a))$~\label{fig:tree_fo}. 
Rectangles represent embeddings and rounded squares neural networks.}
\label{fig:tnn}
\end{figure}
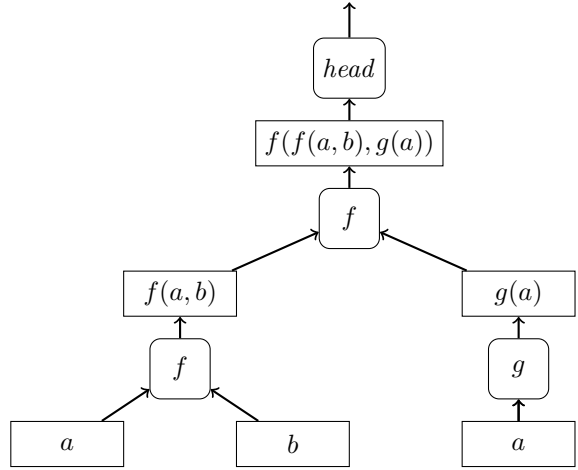

Let $\mathbb{O}$ be a set of operators and $\mathbb{T}_\mathbb{O}$ be the set 
of all terms than can be constructed 
from  $\mathbb{O}$.
A tree neural network (TNN) is a machine learning 
model designed to approximate functions from $\mathbb{T}_\mathbb{O} \mapsto 
\mathbb{R}^n$. We define first the structure of the tree neural network and 
then show how to compute with it. An example is given in Figure~\ref{fig:tnn}).

\paragraph{Definition}(Tree neural network)\\
We define a tree neural network to be a set of feed-forward 
neural networks with n $layers$ and a $tanh$  
activation function for each layer. 
There is one network for each operator $f$ noted $N(f)$ and one for the head 
noted $H$. 
The neural network operator of a function with arity $a$ is to learn a function 
from  $\mathbb{R}^{a \times d}$ to $\mathbb{R}^d$ where $d$ is the dimension
of the embedding space.
And the head network is to approximate a function from $\mathbb{R}^d$ to
$\mathbb{R}^n$.
As an optimization for a operator $f$ with arity 0, $N(f)$ is defined to be
a vector of weights in the embedding space $\mathbb{R}^d$ since multiple layers 
are not needed for learning a constant function.

\paragraph{Computation of the Embeddings}
Given a TNN, we can now define recursively an embedding function $E: 
\mathbb{T}_\mathbb{O} \mapsto \mathbb{R}^d$ by 
$E(f(t_1,\ldots,t_a)=_\mathit{def} \mathit{N}(f)(E(t_1),\ldots,E(t_a))$
This function produces an 
internal representation of the terms to be later processed by the head network.
This internal representation is often called a \textit{thought vector}.

\paragraph{Computation of the Output}
The head network interprets the internal representation and makes the
last computations towards the expected result. In particular, it reduces the 
embedding dimension $d$ to the dimension of the output $n$.
The application of a TNN on a term $t$ gives the result $H(E(t))$.
It is possible to learn a different objective by replacing only the head of the 
network. We use this to our advantage in the reinforcement learning experiments 
where we have the double objective of predicting a policy and a value (see 
Section~\ref{sec:drl}). 

\paragraph{Higher-Order Terms}
A \holfour term can be encoded into a first-order term (i.e. a labeled tree) in 
two steps. First, the function applications $f\ x$ are rewritten to 
$\mathit{apply}(f,x)$ introducing an explicit $\mathit{apply}$ operator.
Second, the lambda terms $\lambda x. t$ are substituted with 
$lam(x,t)$ using the additional operator $lam$. 
For tasks performed on \holfour terms that are 
essentially first-order and these encodings are not required.

\section{Deep Reinforcement Learning}\label{sec:drl}
When possible, the deep reinforcement learning 
approach~\cite{Sutton:1998:IRL:551283} is preferable to a 
supervised learning approach for two main reasons. First, an oracle is not 
required. This means that the algorithm is more general as it does not require 
a specific oracle for each task and can even learn a task for which nobody 
knows a good solution. 
Secondly by decomposing the problem in many steps, the trace of the computation 
becomes visible which is particularly important if one wants a justification 
for the final result.

We present here our methodology to achieve deep reinforcement learning through 
self-supervised learning. It consists of two phases. During the exploration 
phase, a search tree is built by the MCTS algorithm from a prior value and 
prior policy. These priors are updated by looking at the consequence of each 
action. During the learning phase, a new TNN learns to replicate this 
improved policy and value. This new TNN will be used to guide the next 
exploration phase. One iteration of the reinforcement learning loop is called a 
\textit{generation}. In the later part of the paper, we refer to the 
application of the full reinforcement learning loop (alternation of learning 
phases and exploration phases) as the training of the 
TNN. Once trained, we judge the performance of the TNN during a final 
evaluation (Section~\ref{sec:result}). 

Before describing both phases, we define 
what a search problem is, what the 
policy and value for a search algorithm are and how they can be combined 
to guide the search using MCTS.

\subsection{Specification of a Search Problem}~\label{sec:abs_spec}
Any task that can be solved in a series of steps with decision points at 
each step can be used to construct search problems. For example, theorem 
proving is considered a search task by construction. Other para-proving tasks 
are harder to view in such a light, such as: programming, conjecturing, 
making definitions, re-factoring. 
Here, a search task is described as a single-player perfect information game.

\paragraph{Definition}(Search problem)\\
A search problem is an oriented graph.\\
The nodes are a set of states $ \mathbb{S}$ with particular labels for:
a starting state $s_0\in \mathbb{S}$,
a subset of winning states $\mathbb{W} \subset \mathbb{S}$,
and a subset of losing states $\mathbb{L} \subset \mathbb{S}$.\\
The oriented edges are labeled by a finite set of moves $\mathbb{M}$. The 
transition function $T: \mathbb{S} \times \mathbb{M} 
\mapsto \mathbb{S}$ returns the state reached by making a move from the input 
state. To solve the problem, an algorithm needs to find a path $p$ from the 
starting 
state to a winning state avoiding the losing states.
An end state is a state that is either winning or losing.

\paragraph{Policy and Value}
A policy $P$ is a function from $\mathbb{S}$ to $[0,1]^ {\mathit{cardinal} 
(\mathbb{M})}$ that assigns to each state $s$ a real number for each move.
It is intended to be a probability which indicates the percentage of 
times each move should be explored at each state. An impossible move is given a 
policy score of $0$ and thus is never selected during MCTS.

A value $V$ is a function from $\mathbb{S}$ to the interval $[0,1]$. 
$V(s)$ is used as an estimate of how likely the search 
algorithm is to complete the task. Therefore, the function needs to respect 
these additional constraints: a value of $1$ for winning states, and a value of 
$0$ for losing states.

\subsection{Monte Carlo Tree Search}
An in-depth explanation of the Monte Carlo Tree Search algorithm is given in 
\cite{montecarlo}. This search algorithm strikes a good balance between 
exploration of uncertain paths and exploitation of path leading to states with 
good value $V$.
The algorithm was recently improved in~\cite{silver2017mastering}. 
The estimation of the value $V$, which used to be approximated by the 
proportion of random walks (or roll-outs) to a winning state, is now returned 
by a deep neural network. Explicitly, this means that we do not perform 
roll-outs during the node extension steps and instead the learned value 
$V$ is used to provide the rewards.

A search problem is explored by the MCTS algorithm with the help of a prior 
policy $P$ and a prior value $V$. 
The algorithm starts from an initial tree 
(that we call a \textit{root tree}) containing an initial state and proceeds to 
gradually build a search tree. 
Each iteration of the MCTS loop can be decomposed into three main components: 
node selection, node extension and backup. 
One step of this loop usually creates a new node in the search tree unless the 
node selection reached an end state (winning or losing). 
The search is stopped after a fixed number of iterations of the loop called the 
\textit{number of simulations} or after a fixed time limit.
The node selection process is guided by the PUCT 
formula~\cite{DBLP:conf/pkdd/AugerCT13}. We set its \textit{exploration 
coefficient} to 2.0 in our experiments. 
The rewards are calculated as usual with
winning state given a reward of 1 and losing states a reward of 0 and the 
rewards for other states are given by the value $V$.

\subsection{Exploration Phase}
During the exploration phase of the reinforcement learning loop, a full attempt 
at a solution will rely on multiple calls to the MCTS algorithm.
The first call is performed starting from a root tree containing 
the starting state. After an application of the MCTS algorithm, the constructed 
tree is used to decide which move to choose. The application of this move to
the starting state is a \textit{big step}.
The MCTS algorithm is then restarted on a root tree containing the resulting 
state.
This procedure is repeated until a big step results in an end state or
after the number of big steps exceeds a fixed bound.
This bound is fixed to be twice the size of an existing solution found during 
problem generation (Section~\ref{sec:tasks}).
An attempt is successful if it ends in a winning state.

The decision which big steps to make is taken from the number of visits
for each child of the root. During the exploration phase, 
the move with the highest number of visits is chosen. 
To encourage exploration during training, a noise
is added to the prior policy of the root node. We draw noise from a 
uniform distribution and thus do not have to choose the parameter of the 
Dirichlet distribution as in~\cite{silver2017mastering}.

\paragraph{Remark}
The upper bound is also used to limit the depth of the MCTS calls 
during training. Since it gives our prover some problem specific 
knowledge about the limits of the search space and also prevents it 
from finding longer solutions, the bound is not enforced 
during the final evaluations on the testing sets.

\paragraph{Problem Selection}
It is computationally expensive to make attempts on all 2000 problems from the 
training set (see Section~\ref{sec:datasets}).
That is why we select a subset of 200 problems. As we aim to make 
the dataset of examples balanced, we select 100 positive problems and 100 
negative problems. A problem is positive if it was attempted previously and 
solved in the last attempt and negative otherwise. 
On top of that, we try to focus the selection on problems away from problems 
that are too hard or too easy. To estimate how easy and hard a problem is we
look at the list of successes and failures of the algorithm on this problem.
We count the number of successes (respectively failures) in a row starting from
the end of the list on a positive (respectively negative) problem. 
The inverse of this number is higher for a problem that has just shifted from 
negative to positive or conversely.
This score is normalized into a probability distribution across all the 
positive (respectively negative) 
examples. The selected positive (respectively negative) problems are drawn from 
this distribution.

\subsection{Learning Phase}
After each big step of the exploration phase, we collect an example consisting 
of an input term representing a search state associated with an improved policy 
and an improved value for this search state.
More precisely, the example is constructed from statistics of the root of the 
search tree (the root changes after each big step). The input of the example 
is the term representing the state of the 
root. The improved policy for this example is computed by dividing the number 
of visits for each child by the total number of visits for all children.
The improved value is the average of the values of all the nodes in the tree 
with the value of an end state counted as many times as it has been visited.
The new example is added to a dataset of training examples. 
This dataset is then used to train the TNNs in future generations.
When the number of examples in the dataset reaches $x$, older 
examples are discarded whenever newer examples are added so that the number of 
examples never exceeds $x$. This number is called the \textit{size of the 
window} and is set to 200,000 in our experiments.

The TNN learns to imitate the collected policy and value 
examples by following the batch gradient descent 
algorithm~\cite{DBLP:conf/kdd/LiZCS14}. In our implementation, we 
rely on a mean square error loss and update the weights using backpropagation. 
Furthermore, $P$ and $V$ are learned simultaneously by a 
double-headed TNN. In other words, the learned term embeddings are shared
during the learning of the two objectives.

\paragraph{Remark}
The states (or their term representation) in the policy and value dataset
differs from the dataset of problems. Indeed, 
problems are starting states whereas policies and values are extracted for 
all intermediate states covered by the big steps.

\section{Specification of the Tasks}~\label{sec:tasks}
In this section, we specify the two synthesis tasks that are used for 
experiments with 
our framework. We describe how to express the search problems and provide 
optimizations to facilitate the learning.

\subsection{Synthesis of Combinators}\label{sec:combin}
The aim of this task is to find a SK-combinator $c$ 
such that $c\ x_1\ x_2\ \ldots\ x_n \rightarrow 
h'[x_1,x_2,\ldots,x_n]$
where $h'$ is a higher-order term only composed of the functions 
$x_1,x_2,\ldots,x_n$ and $\rightarrow$ is rewrite relation given by
the term rewrite system $\lbrace S x y z \rightarrow (x z) (y 
z),\ K x y \rightarrow x \rbrace$,
A combinator is by definition said to normalize if has a normal form. This 
unique normal form can always be obtained using the left-outermost strategy.
Since having a solution $w$ to the problem implies that the combinator $w\ x_1\ 
x_2\ \ldots\ x_n$ normalizes and therefore $w$ normalizes too and its normal 
form $w'$ is also a solution. Thus, we can limit the search to SK-combinators 
in normal form.

To synthesize combinators we introduce a meta-variable $X$ as a place holder at 
positions of to-be-constructed subterms.
The starting state is a couple of the partially synthesized term $X$ and 
$h'[x_1,x_2,\ldots,x_n]$.
A move consists oFf applying one the 
following five rewrite rules to the first occurrence of $X$ in the first 
component of the state:
\[X \rightarrow S,\ X \rightarrow S X,\ X \rightarrow S X X,\
  X \rightarrow K,\ X \rightarrow K X\]
This system is exactly producing all the combinators in normal forms. It 
prevents the creation of any redex by limiting the number of arguments of $S$ 
and $K$.
A state is winning if the synthesized witness $w$ applies to $x_1\ 
x_2\ \ldots\ x_n$ rewrites to the normal form $h'[x_1,x_2,\ldots,x_n]$, which 
is determined in practice by applying the left-outermost rewrite strategy. We 
remove occurrences of $X$ in $w$ before testing for the winning condition.
A state is losing if the synthesized witness $w$ does not contain $X$ and is 
not a solution.

%\paragraph{Example}
%The synthesis of $S (K S) K$ is achieved by the rewrite sequence
%$X \rightarrow S X X \rightarrow S (K X) X \rightarrow S (K S) X \rightarrow S 
%(K S) K$ . The combinators $S$, $S K$ and $S (K S)$ are tested as potential 
%solutions along the way.

\paragraph{TNN Representation of the State}
To make its decision between the five moves and evaluate the current state, 
the TNN receives components of the state as first-order 
terms by tagging each operator with its local arity and merge their embeddings 
using a concatenation network. 
For instance, the combinator $S (K S) K$ is encoded as $s_2(k_1(s_0),k_0)$. 
Compared with the apply encoding, 
the advantage is that we can represent each combinator with arity greater than 
zero as a neural network instead of an embedding in our TNN architecture.
The drawback is that the learning is split between operators of 
different arities.

\paragraph{Complexity}
For synthesizing a solution of size 20, an upper bound for the search space of 
our algorithm is $5^{20} \approx 9.5 \times 10 ^ {13}$ as there are 5 possible 
moves.
Since no moves are allowed from end states, there are more precisely at
most $4.9 \times 10^{11}$ states with a partially synthesized combinator of 
size less than 20.

\subsection{Synthesis of Diophantine Equations}\label{sec:dioph}

The aim of this task is given a particular set $S$ to find a polynomial whose 
Diophantine set is equal to $S$. 
In the following description and for the experiments, 
we limit the range of polynomials to the ones with one parameter $k$ and three 
existential variables $x,y,z$. The Diophantine set $D(p)$ of the polynomial $p$
is defined by $D(p) =_{\mathit{def}} \lbrace k\ |\ \exists xyz.\ p(k,x,y,z) = 0 
\rbrace$.
To make the computation of $D(p)$ tractable for any polynomial, the domain of 
interpretation of the variables and operators is changed to 
$\mathbb{Z}/16\mathbb{Z}$.

The polynomials considered for synthesis are normalized polynomials.
They are expressed as a sum of monomials. 
Internally, each monomial is represented as a list of integers. The first 
element of the list is the coefficient, and the remaining elements are the 
exponents of the variables $k$,$x$,$y$ and $z$.
A polynomial is a list of monomials and its size is the sum of the length of 
its monomials (represented as lists). For instance, the monomial
$k^2x^3 + 2y^4$ is internally represented as $[[1, 2, 3], [2, 0, 0, 4]]$.
Since multiplication is associative 
and commutative (AC), variables are ordered alphabetically 
in the monomial. Since addition is AC, monomials are sorted by comparing
their list of exponents using the lexicographic order.

Our starting state consists of the empty polynomial and an enumeration $S$.
To synthesize a polynomial we rely on two types of moves. The first one
is to start constructing the next monomial by choosing its coefficient. The 
second one is to choose an exponent of the next variable in the current 
monomial. In practice, we limit the maximum value of an exponent to 4 and
the number of monomials per polynomial to 5.
A state $(w,S)$ is winning if the polynomial $w$ defines the set $S$.

\paragraph{TNN Representation of the State}
Computing the embedding of a state $(w,S)$ relies on a neural network operator 
for merging the embeddings of $w$ and $S$.
The set $S$ is encoded as a list of 16 real numbers. 
The $i^{th}$ element of this list is 1 if $i\in S$ and -1 otherwise.
The polynomial $w$ is represented using a tree structure with operators $+$ and 
$\times$ and we have learnable embeddings for
each variable and each coefficient. To compress the representation further,
variables with their exponent (e.g. $x^3$) are represented as a single 
learnable embedding instead of a tree.

\paragraph{Complexity}
An approximation measure for the size of the search space is the
number of polynomials in normal form. This number is $\frac{(16 \times 5^4) ^ 
5}{5!} \approx 8.3 * 10 ^ {17}$. 

\section{Datasets}\label{sec:datasets}
In all our tasks, our algorithms require a training set in order to learn the 
task at hand. In a reinforcement learning setting, a training problem 
does not come with its solution as in supervised learning, thus
problem-solving knowledge cannot be obtained by memorization and has to be 
acquired through search. Still, we also create an independent testing set to 
further estimate the generalization abilities of the 
algorithm on problems not seen during training. 
Even in the context of reinforcement learning, the ability of TNNs to learn a 
task is heavily influenced by the quality of the training examples.
The following objectives should guide the generation of the training set: a 
large and diverse enough set of input terms,
a uniform distribution of output classes and a gradual increase in difficulty.
%The number of examples generated should be proportional to the set of 
%operators in our training examples as they influence the number of parameters 
%of the TNN.That is why we try use a minimal number of operators to represent 
%the 
%terms. 

For both tasks, problems are generated iteratively in the same way.
At the start, the set of problems $\mathbb{P}$ is empty.
At each step, a random witness $w$ (polynomial or combinator) is produced and 
we compute its image $f(w)$. 
If the image does not have the desired form, then $\mathbb{P}$ remains 
unchanged. 
If the image does not exist in $\mathbb{P}$, we 
add the problem represented by $f(w)$ and its solution $w$ to the problem set.
If the image already exists in the set and the witness is smaller
than the previous one for this image, then we replace the previous solution
by the new one. If it is bigger, then $\mathbb{P}$ remains unchanged. 
We repeat this process until we have 2200 distinct problems. 
This set of problems is split randomly into a training set of 2000 problems and
testing set of 200 problems. We use the generated solutions to estimate the 
difficulty of the problems and bound the number of big steps during training. 
The generated solutions are not revealed to any other part of the algorithm. In 
particular, no information about these solutions is used during the final 
evaluation on the test set.

To generate a random combinator, we pick randomly a size between 1 and
20 and then draw uniformly at random from the set of normal form SK-combinators
of that size. Generating this set becomes too computationally expensive for 
a size greater than 10, thus we rely on a top-down generation that exactly 
simulates the process.  It works by selecting the top operator and the size of 
its arguments according to their frequencies which can be computed much more 
efficiently.
To generate a random polynomial, we first select a number of monomials in 
[|1,5|]. Then for each monomial, we choose a number of variables in [|0,4|], a 
leading coefficient in [|1,15|] and an exponent in [|0,4|] for each variable.

\begin{figure}[]
\begin{tikzpicture}
\begin{axis}[
  legend style={anchor=north east, at={(0.95,0.95)}},
  width=\textwidth,
  height=0.4*\textwidth,xmin=1, xmax=22,
  ymin=0, ymax=300,
  cycle list name=rw2
  ]
\addplot table[x=size, y=pb] {combin_dis};
\addplot table[x=size, y=pb] {poly_dis};
\legend{combinators,polynomials}
\end{axis}
\end{tikzpicture}
\caption{\label{fig:dis} Number (y) of problems a generated solution of size 
(x)}
\end{figure}
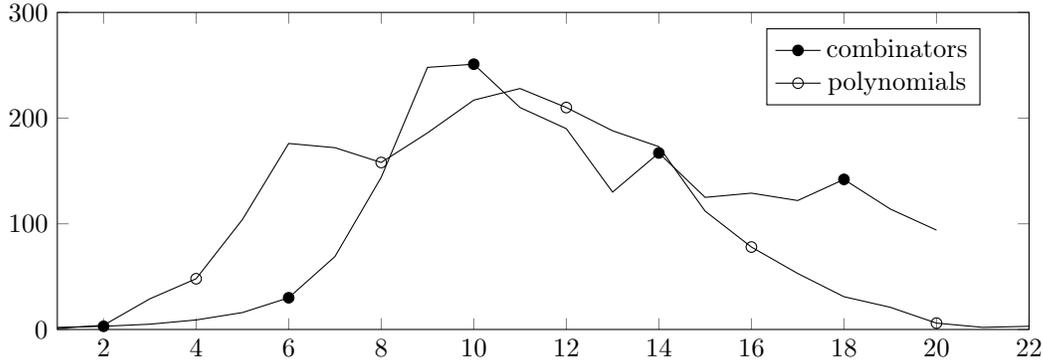

Figure~\ref{fig:dis} shows the distributions of the problems according to their 
difficulty. There, the size of the smallest generated 
solution is used as a measure of difficulty for each problem.

\pgfplotscreateplotcyclelist{rw2}
{solid, mark repeat = 4, mark phase = 2, mark = *, black\\
 solid, mark repeat = 4, mark phase = 4, mark = o, black\\}

\section{Results}\label{sec:result}
The following experiments demonstrate how the reinforcement learning 
framework is able to gradually learn each task by recursive 
self-improvement. We first analyze the progress made during training,
Then, we compare our method with alternatives during evaluation.
Finally, we analyze the distribution of solutions produced to gain some 
insight into what has been learned.

\paragraph{Replicability}
The code for the framework and the experiments are available in the
repository for 
\holfour\footnote{\url{https://github.com/HOL-Theorem-Prover/HOL}}.
Although the code is available on top of the master branch, to be able to import
the provided HOL4 datasets of combinator problems, one needs to switch to
the commit bcd916d1251cced25f45c90e316021d0fd8818e9 as the format for exporting 
terms was changed.

The specification of each task is implemented in the 
\textsf{examples/AI\_tasks}
directory. The underlying framework shared by both tasks is located in the
\textsf{src/AI} directory. 
The file \textsf{examples/AI\_tasks/README} explains how to reproduce the 
experiments. 
The datasets can be downloaded from our repository
\footnote{\url{https://github.com/barakeel/synthesis_datasets}}.

\paragraph{Parameters}
The parameters of the TNN used during our experiments are an embedding 
dimension of size 16, one fully connected
layer per operator and two fully connected layers for the policy head and the 
value 
head. The schedule of the learning phase consists of 10 epochs on a maximum 
of 200,000 examples and a learning rate of 0.02.

\subsection{Combinators}
Experiments on combinators rely on an instance of the MCTS 
algorithm given by the specification from Section~\ref{sec:combin}
noted MCTS$_{\mathit{combin}}$.

\begin{figure}[]
\begin{tikzpicture}
\begin{axis}[
  legend style={anchor=north east, at={(0.95,0.4)}},
  width=\textwidth,
  height=0.5*\textwidth,xmin=0, xmax=318,
  ymin=0, ymax=1700,
  cycle list name=rw
  ]
\addplot table[x=gen, y=sol] {combin_graph};
\addplot table[x=gen, y=exp] {combin_graph};
\legend{at least once, expectancy}
\end{axis}
\end{tikzpicture}
\caption{\label{fig:combin} Number y of training problems solved after 
generation x}
\end{figure}
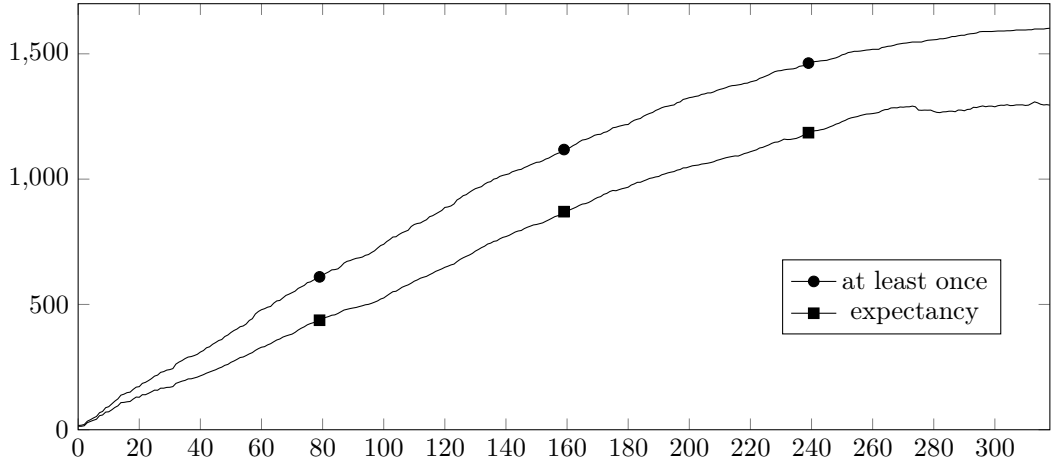

\paragraph{Training}
Figure~\ref{fig:combin} shows the number of problems solved at least once by
MCTS$_{\mathit{combin}}$ (run with multiple big steps) over 318 
generations. 
To give a better view of the progress of the algorithm, we also show the
number of problems that we expect the algorithm to be able to solve. This 
number is obtained by
computing the frequency at which it solves each problem on the last five tries 
and summing up the frequencies across all problems.
This graph shows that little improvement occurs after generation 270.
The discrepancy between the two lines also indicates that our TNN is not able
to memorize perfectly the previously discovered proofs. One solution to these 
issues would be to increase the dimension of the embeddings. The trade-off 
is that it will slow down quadratically the speed of inference and may 
lead to a weaker generalization ability if applied without proper 
regularization.
All but 100 examples are attempted before generation 34. At this point, only 135
solutions are found. This is to be compared to the total of 1599 solutions 
found by the end of the training.  
The oldest examples retrieved from searches are discarded after generation 64 
as their number surpasses 200,000.
After each generation, we save the weights of the trained TNN. In the following
evaluation, we use the weights of the TNN from generation 318.

\paragraph{Evaluation}
\begin{table}[]
\centering
\begin{tabular}{llcc}
\toprule 
Prover & Strategy & Train (2000) & Test (200)\\
\midrule
\eprover 2.4 \cite{eprover}& auto & 38.80 & 36.0 \\
         & auto-schedule & 50.35 & 48.5\\  
\vampire 4.2.2 \cite{DBLP:conf/cav/KovacsV13} & default & \phantom{0}4.15 & 
\phantom{0}3.5\\
         & CASC mode & 63.45 & 62.0  \\
MCTS$_{\mathit{combin}}$ & uniform & 27.65 & 27.0 \\
 & TNN-guided & 72.70 & 65.0 \\
\bottomrule
\end{tabular}
\caption{Percentage of problems solved within 60 seconds}\label{tab:combin}
\end{table}

During the final evaluation, we run each prover on one CPU for 60 seconds on 
each problem. Their success rates are presented in 
Table~\ref{tab:combin}. The MCTS$_{\mathit{combin}}$ algorithm is run without 
noise and for as many simulations as the time limit allows. In particular, we 
do not apply any big steps (non-backtrackable steps) as they were mainly 
introduced to produce training examples for states appearing deeper in the 
search. By comparing the results of MCTS$_{\mathit{combin}}$ on the training 
and testing set, we observe that it generalizes well although not perfectly to 
unseen data.
Our machine learning guided can then be compared with the uniform strategy 
(approximating breadth-first search) where each branch is explored with the 
same probability. Since the uniform strategy does not call the TNN, it can 
perform on average more than twice as many simulations (414,306 vs 196,000) in 
60 seconds. Even with this advantage, the trained algorithm outperforms it 
significantly.

To compare our algorithm with state-of-the-art automated theorem provers, the
synthesis problem is stated using an existential first-order formula for the 
conjecture. As an 
example, the problem for 
synthesizing a combinator equal to the head normal form $\lambda v_1v_2v_3.v_3$
can be expressed in the TPTP~\cite{tptp} format (an input format for ATPs) as:
\begin{verbatim}
fof(axS,axiom, ![X, Y, Z]: (a(a(a(s,X),Y),Z) = a(a(X,Z),a(Y,Z)))).
fof(axK,axiom, ![X, Y]: (a(a(k,X),Y) = X)).
fof(conjecture,conjecture, ?[Vc]: ![V1, V2, V3]: (a(a(a(Vc,V1),V2),V3) = V3)).
\end{verbatim}

Even using their more advanced set of strategies (auto-schedule for \eprover 
and CASC mode for Vampire), the success rates of ATPs are less than the trained 
MCTS$_{\mathit{combin}}$ on both the training and testing sets.
It is worth mentioning that the ATPs and our systems work in a very different
way on the synthesis tasks. In our system, there is a synthesis part guided by 
our TNN and a checking part performed by a 
deterministic normalization algorithm. 
In contrast, the ATPs are searching for a proof by applying the rules of their 
calculus. They essentially deduce intermediate lemmas (clauses) to
obtain the proof. As a comparison, the number of generated clauses by \eprover
on average on this dataset in 60 seconds is about three million.
 One advantage of the approaches of ATP is that they might be 
able to split the synthesis problem is smaller ones by finding an independent 
part of the head normal form. The trade-off is that with more actions available,
smarter strategies are needed to reduce the search space. We believe that 
combining both approaches, i.e. learning to guide searches on the ATP calculus, 
would certainly lead to further gains.
Finally, our algorithm naturally provides a 
synthesized witness but it may be harder to extract such witness from an ATP 
proof. Therefore, we can now analyze the witnesses provided as solutions during 
the final evaluation.

%maybe cut out the computation of the normal form.
\paragraph{Examples}
The $\lambda$-abstraction $\lambda fxy.fyx$ gives a semantic description of the 
$C$ combinator commonly used in functional programming.
The solution $S (S (K S) (S (K K) S)) (K K)$ proposed by our algorithm is 
not equivalent to the solution $S (B B S) (K K)$ given by 
Schönfinkel~\cite{schonfinkel1924bausteine}
with $B = S (K S) K$.  Its normal form 
$S (S (K (S (K S) K)) S) (K K)$ is different.
The witness synthesized by MCTS$_{\mathit{combin}}$ for
$\lambda xyz.\ x y (x y) (y (x y) (x y (y (x y)))) z$
is the largest among the solutions. This combinator is
$S (S (S (K (S (S (K S)))) K) S) (S (S (S (S K K))))$.

\paragraph{Analysis of the Solutions}
In Table~\ref{tab:combin_occ}, we observe patterns in the subterms of 
the solutions by measuring how frequently they occur. This only gives us a very 
rough understanding of what the TNN has learned since the decisions made 
by the TNN are context-dependent.
The combinator $S$ occurs about twice as often as $K$. More interestingly,
the combinator $S K K$ occurs 585 times whereas the combinator $S K S$ which 
as the same effect (i.e. the identity) only 39 times. Having strong preferences 
between equivalent choices is beneficial as it avoids duplicating searches. 
However, how this preference was acquired is still to be determined.
The combinator $B$ appears in fourth position among the 40 combinators in 
normal form of size four showing its importance as a building block for 
combinator synthesis.

\begin{table}[]
\centering\small
\begin{tabular}{lccccccc}
\toprule
Subterm & $S$ & $K$ & $S S$ & $S K$ & $K S$ & $S (K S)$ & $S K K$ 
\\
Occurrences & 11187 & 5114 & 1883 & 1082 & 710 & 709 & 585\\
\midrule
Subterm & $S S K$ & $K (S S)$ & $S (K (S S))$ & $S (S K K)$ & $S (S S)$ & $S S 
(S K)$ & $S (S S K)$ \\
Occurrences & 370 & 305 & 305 & 273 & 245 & 204 & 183 \\
\midrule
Subterm  & $K K$ & $S (K K)$ & $S (K S) K$ & $S (S (S K K))$ & $S (K S) S$ & 
$S (S (K S) K)$ & \\
Occurrences & 173 & 165 & 153 & 144 & 
138 & 135 & \\
\bottomrule
\end{tabular}
\caption{Number of occurrences of the 20 most frequent subterms 
that are part of the 1584 combinator solutions of the 2200 combinator problems}
\label{tab:combin_occ}
\end{table}

\subsection{Diophantine Equations}
Experiments on Diophantine equations rely on an instance of the MCTS 
algorithm given by the specification from Section~\ref{sec:dioph}
noted MCTS$_{\mathit{dioph}}$.

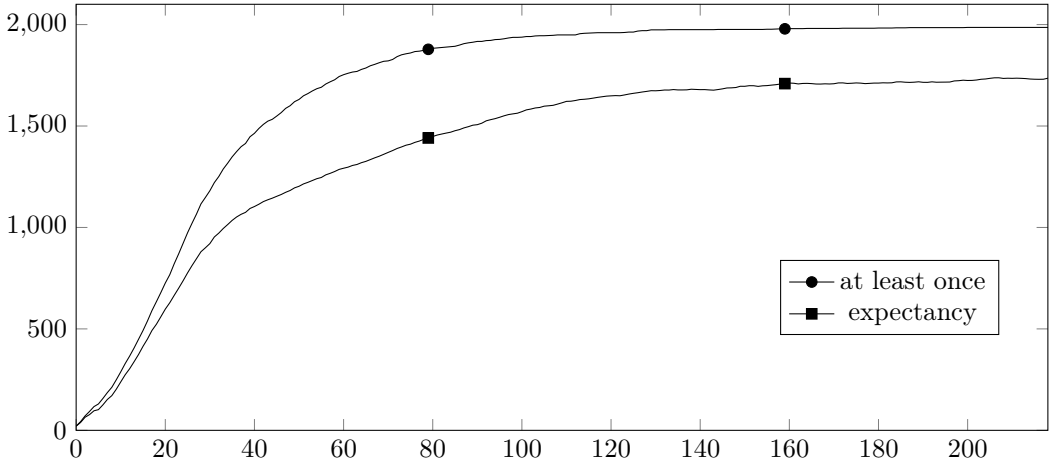
\begin{figure}[]
\begin{tikzpicture}
\begin{axis}[
  legend style={anchor=north east, at={(0.95,0.4)}},
  width=\textwidth,
  height=0.5*\textwidth,xmin=0, xmax=218,
  ymin=0, ymax=2100,
  cycle list name=rw
  ]
\addplot table[x=gen, y=sol] {dioph_graph};
\addplot table[x=gen, y=exp] {dioph_graph};
\legend{at least once, expectancy}
\end{axis}
\end{tikzpicture}
\caption{\label{fig:dioph} Number y of problems solved after generation x}
\end{figure}

\paragraph{Training}
The evolution of the success rate during training is presented in 
Figure~\ref{fig:dioph}. At the end of the training, the number of problems
solved at least once is 1986 out of 2000. However, the expectancy is quite 
lower showing the same issue in the memorization ability of the TNN as for 
the combinators. This might be solved by increasing the dimension of the 
embeddings or by improving the network architecture. The number of examples 
reaches 200,000 at generation 84 later than in the combinator experiments, 
indicating that the polynomials synthesized are shorter. In general, the fact 
that we obtain a much higher success rate in this experiment indicates that a 
higher branching factor but shallower searches suit our algorithm better.
In the final evaluation, we use the weights of the TNN from generation 197.

\paragraph{Evaluation}
The results of the final evaluation presented in Table~\ref{sec:dioph} show
a drastic difference between the uniform strategy and the TNN-guided one. 
Thus, this task produces many patterns that the TNN can 
detect. This observation is reinforced by the fact that the uniform algorithm 
performs on average 3 to 4 times more simulations in 60 seconds (269,193 vs 
79,976) than the TNN-guided one. 
To create a stronger competitor, we handcraft an evaluation function
reflecting our intuition about the problem. An obvious heuristic is to guide 
the algorithm by how many of the elements of $[|0,15|]$ the current polynomial 
correctly classifies as a member of the targeted set or not. This number is 
then 
divided by 16 to give an evaluation function 
between 0 and 1. The results are displayed in the heuristic row in 
Table~\ref{sec:dioph}. Surprisingly, the heuristically guided search performed
worse than the uniform strategy mainly due to an overhead cost for computing 
the evaluation resulting in 70,233 simulations (on average) per MCTS calls.
We do not know of any higher-order theorem prover with support for arithmetic.
Finding an encoding of the arithmetic operations and/or the higher-order 
features, which makes the problems tractable for a targeted ATP, is
non-trivial and is not investigated in this paper. That is why we did not 
compare our results with ATPs on Diophantine equations.

\begin{table}[]
\centering
\begin{tabular}{lcc}
\toprule
Strategy & Train (2000) & Test (200)\\
\midrule
uniform & \phantom{0}3.70 & \phantom{0}4.0 \\
heuristic & \phantom{0}3.05 & \phantom{0}0.5 \\
TNN-guided & 77.15 & 78.5\\
\bottomrule
\end{tabular}
\caption{Percentage of problems solved in 60 seconds by
MCTS$_{\mathit{Dioph}}$}\label{tab:dioph}
\end{table}

\paragraph{Example}
The solution for the set $\lbrace 0, 1, 3, 4, 5, 9, 11, 12, 13 \rbrace$ is the 
largest among all solutions. The algorithm generated the polynomial 
$y^2 + 12x^4 + 7k + 7k^2x^2y^2 + 7k^2x^2y^2z^2$.

\paragraph{Analysis of the Solutions}

\begin{table}[]
\centering\small
\begin{tabular}{lcccccccccc}
\toprule
Monomial & $7k$ & $7k^2x^2$ & $7k^2x^2y^2$ & $7k^2$ & $7kx^2$ & $7$ & $14$ 
 &$7k^4$ & $7k^4x^2$ & $12$\\
Occurences & 780 & 682 & 640 & 387 & 354 & 329 & 218 & 149 & 144 & 143\\
\midrule
Monomial & $8$ & $7k^3$ & $14k$ & $7k^2x^2y^2z^2$ & $4$ & $6$ & 
$7k^4x^2y^2$ 
& $10$ & $10k$ & $2$ \\
Occurences & 118 & 111 & 106 & 92 & 90 & 74 & 72 & 69 & 68 & 61\\
\bottomrule
\end{tabular}
\caption{Number of occurrences of the 20 most frequent monomials 
that are part of the 1700 polynomial solutions of the 2200 problems on 
Diophantine equations}
\label{tab:dioph_freq}
\end{table}

The most frequent monomials appearing as part of a polynomial solution are
shown in Table~\ref{tab:dioph_freq}. The coefficients of those monomials are 
either even or 7. Only the monomials with coefficient 7 contain
an existential variable $x$,$y$ or $z$. The exponent of an existential variable
is always 2. Following these three patterns exactly limits the search space 
but might make some problems unsolvable. 
The two previous examples contain monomials outside of these patterns. 
This shows that the algorithm is able to deviate from them when necessary.

\section{Verification}
A final check to the correctness of our algorithm can be 
made by verifying the solutions produced during evaluation. 
The verification consists of producing HOL4 theorems stating that each 
witness $w$ has the desired property $\mathit{f}(w) = c$. The function $f$ is 
specified by the task and the image $c$ is given in the problem.
In the following, we present how to construct a verification procedure for an 
arbitrary problem/solution pair for the two tasks. 

\subsection{Combinators}
First, we express the properties of combinators as HOL4 formulas.
There exists an implementation of typed combinators in HOL4 but they cannot be 
used to represent untyped combinators. For instance, the combinator $S 
(S K K) (S K K)$ constructed with HOL4 constants is not well-typed. 
Therefore, we use the apply operator $apply$ of type $\alpha \rightarrow \alpha 
\rightarrow \alpha$ and the type $\alpha$ for representing the type of 
combinators. We note $x.y$ the term $apply(x,y)$.
 Definitions of the free variables $s$ and $k$ representing the 
combinators $S$ and $K$ are added as a set of assumptions to every problem. 
From a witness $w$ and a head normal form $h=\lambda xyz.h'(x,y,z)$,
we prove the theorem:
\[\lbrace \forall xyz.\ ((s.x).y).z = (x.z).(y.z),\ \ \forall 
xy.\ (k.x).y = y \rbrace\ \vdash\ \forall x y z.\ ((w.x).y).z = h'(x,y,z)\]

A call to the tactic \texttt{ASM\_REWRITE\_TAC []} completes the proof.
The function \texttt{COMBIN\_PROVE}, following the described process, 
was used to verify the correctness of the 130 solutions found by 
MCTS$_\mathit{combin}$ on the test set.

\subsection{Diophantine Equations}
Since problems on Diophantine equations are about sets of natural numbers 
described by using arithmetic operations, we rely on \holfour constants and 
functions from the theories \texttt{num}, \texttt{arithmetic} and 
\texttt{pred\_set} to express that the solution satisfies the problem.
We verify that $D(w)$ (the Diophantine set described by the polynomial $w$)
is the enumeration $\lbrace a_1,a_2,\ldots,a_i \rbrace$. The equality between 
the 
two sets can be expressed in \holfour as:
 \[\lbrace k\ |\ k < 16 \wedge 
  \exists xyz.\ w(k,x_{16},y_{16},z_{16}) 
\ \mathit{mod}\ 16 = 
  0 \rbrace = \lbrace a_1,a_2,\ldots,a_i 
  \rbrace\] 
For any variable $v$, the shorthand $v_{16}$ stands for 
$v\ \mathit{mod}\ 16$.
All natural numbers are expressed using the standard HOL4 
natural numbers. All variables have the type of HOL4 natural numbers. That 
is why, in order to reason modulo 16, each existential variable $v$ is replaced 
by $v\ \mathit{mod}\ 16$ and the parameter $k$ is bounded by $16$.

\paragraph{}
The proof starts by considering two predicates $P$ and $Q$ defined by:
\begin{align*}
P &=_{\mathit{def}} 
  \lambda k.\ (\exists xyz.\ w(k,x_{16},y_{16},z_{16}) = 0),\ \
Q =_{\mathit{def}}
  \lambda k.\ (k = a_1 \vee  k = a_2 \vee ... \vee k = a_i)
\end{align*}
To verify that these predicates are equivalent on a particular element $k \in 
[|0,15|]$, we distinguish between two cases. Either the Diophantine equation 
has a solution and both predicates are true, or it does 
not admit a solution and both predicates are false. In both cases, ground 
equations are proven using \texttt{EQT\_ELIM o EVAL} where \texttt{EVAL}  
is a rule (a function that returns a theorem) for evaluating ground expressions.

\paragraph{Positive case}
Let us assume that $k\in \lbrace a_1,a_2,\ldots,a_i \rbrace$.
To prove $P\ k$, we need to prove that $\exists xyz. w(k,x_{16},y_{16},z_{16})
\ \mathit{mod}\ 16 = 0$. Therefore, we search for the triple 
$(a,b,c) \in [|0,15|]^3$ for which the ground equation holds. We call the
resulting theorem $\mathit{thm\_abc}$.
The goal $P\ k$ can then be closed by applying the tactic:
\begin{lstlisting}[language=SML]
EXISTS_TAC a THEN EXISTS_TAC b THEN EXISTS_TAC c THEN ACCEPT_TAC thm_abc
\end{lstlisting}
$Q\ k$ is proven by beta-reduction and a call to a decision procedure for 
ground arithmetic:
\begin{lstlisting}[language=SML]
CONV_TAC (TOP_DEPTH_CONV BETA_CONV) THEN DECIDE_TAC
\end{lstlisting}
\vspace{-5mm}
\paragraph{Negative case}
Let us now assume that  $k \not \in \lbrace a_1,a_2,\ldots,a_i \rbrace$.
To prove $\neg(P\ k)$, we need to prove that
$\forall xyz. w(k,x_{16},y_{16},z_{16}) \ \mathit{mod}\ 16 \not= 0$.
We deduce the following lemma for a predicate $R$ and a variable $v$:
$R\ 0 \wedge R\ 1 \wedge \ldots \wedge R\ 15\ \vdash\ 
\forall v.\ R\ v_{16}$.
Using this lemma, we can reconstruct the universally quantified theorems 
from the proof all possible instantiations:
\begin{align*}
&\vdash\ w(k,0..15,0..15,0..15) \ \mathit{mod}\ 16 \not= 0\\
&\vdash\ \forall z.\ w(k,0..15,0..15,z_{16}) \ \mathit{mod}\ 16 \not= 0\\
&\vdash\ \forall yz.\ w(k,0..15,y_{16},z_{16}) \ \mathit{mod}\ 16 \not= 0\\
&\vdash\ \forall xyz.\ w(k,x_{16},y_{16},z_{16}) \ \mathit{mod}\ 16 \not= 0
\end{align*}
The notation $\vdash t[0..15]$ is a shorthand for $\vdash t[0] \wedge t[1] 
\wedge \ldots \wedge t[15]$.
The proof of $\neg(Q\ k)$ relies on the same tactic as for proving $Q\ k$ in 
the previous case.
 
\paragraph{}
By combining the positive and negative cases with some simple propositional 
reasoning, we obtain equivalences for 
all $k \in [|0,15|]$ .
From these equivalences, we prove the following lemma about the bounded sets:
   \[(P\ 0 \Leftrightarrow Q\ 0) \wedge \ldots 
   \wedge (P\  15 \Leftrightarrow Q\  15)\ \vdash \  
   \lbrace k\ |\ k < 16 \wedge P\  k \rbrace = 
   \lbrace k\ |\ k < 16 \wedge Q\  k \rbrace \]
The final step is to convert the set defined by $Q$ into
an enumeration by proving the lemma:
\[ \vdash\ \lbrace k\ |\ k < 16 \wedge (\lambda k.\ (k = a_1 \vee  k = a_2 \vee 
... \vee k = a_i))\  k \rbrace = \lbrace a_1,a_2,\ldots,a_i \rbrace\]

The function \texttt{DIOPH\_PROVE} encompassing this process was used to 
verify the correctness of the 157 solutions found by 
MCTS$_\mathit{dioph}$ on the test set.

\section{Related Work}
The related work can be classified into three categories:
automation for solving synthesis tasks,
machine learning guidance inside ATPs, learning assisted reasoning in ITPs.
We present the most promising projects in each category 
separately. Their description shows how they compare to our approach and 
influence our methods.

First, synthesizing SK-combinators using machine learning from their defining 
  property has been attempted in~\cite{DBLP:conf/cade/Fuchs97}. There, they use 
  a genetic algorithm 
  to produce 
  combinators with a fitness function that encompasses various heuristics.
  By comparison, we essentially try to learn this fitness function from 
  previous search attempts.
  The bracket abstraction algorithm developed 
  by Schönfinkel~\cite{schonfinkel1924bausteine} can be 
  used to eliminate a variable and by abstracting all variables achieves 
  SK-combinator synthesis.
  An improvement of the abstraction algorithm using
  families of combinators is proposed in \cite{DBLP:conf/flops/Kiselyov18}.
  A more recent work attempt to solve combinatory logic synthesis problems 
  using a SMT solvers~\cite{DBLP:journals/corr/abs-1908-09481}.
  Matiyasevich proved~\cite{10009422455} that every enumerable set is a 
  Diophantine set.
It is theoretically possible although not practical to extract an algorithm 
from his proof.
  The closest attempt at synthesizing Diophantine equations does not rely on 
  statistical learning and is concerned with the synthesis of polynomials over 
  finite Galois fields~\cite{DBLP:conf/iccad/JabirPM06}.
  If we consider functions to be programs, there is a whole domain of research
  dedicated to program synthesis. Among these, this 
  approach~\cite{DBLP:journals/corr/abs-1911-10244} 
  relies on deep reinforcement learning.

Second, the work that comes closest to achieving our end goal of a competitive 
learning-guided theorem prover is described 
in~\cite{DBLP:conf/nips/KaliszykUMO18}. There, a guided MCTS algorithm is 
trained with reinforcement learning. Its objective is to prove first-order 
formulas from Mizar~\cite{mizar10} problems using a connection-style 
search~\cite{DBLP:journals/jsc/OttenB03}.
The experiments show that gradual improvement stops on the test set after the 
fifth generation. This is probably for two reasons: the small number 
of problems relative to the diversity of the domains considered and 
the inherent limitations of the feature-based predictor~\cite{xgboost} they 
rely on.
Another approach is to modify state-of-the-art ATPs by introducing 
machine-learned heuristics to influence important choice points in their search 
algorithms.
A major project is the development of
\textsf{ENIGMA}~\cite{DBLP:conf/mkm/JakubuvU17,
DBLP:conf/cade/ChvalovskyJ0U19,DBLP:journals/corr/abs-1904-01677}
which guides given clause selection in \eprover~\cite{eprover}. There, a fast 
machine learning model is trained to evaluate clauses from their contribution 
to previous proofs. A significant slowdown detrimental to 
the success rate of \eprover occurs when trying to replace the fast 
predictors by deep neural networks~\cite{DBLP:conf/lpar/LoosISK17}.

Third, our work aims to ultimately bring more automation to ITP users. 
Hammers~\cite{hammers4qed} rely on machine learning guided 
premise selection, translation to first-order and calls to external ATPs
to provide powerful push-button automation. An instance of such a system 
is implemented in \holfour~\cite{tgck-cpp15}. Its performance on induction 
problems is limited by the encoding of the translation.
To solve this issue, the tactical prover 
\tactictoe~\cite{tgckju-lpar17,gkukn-toappear-jar18}, also 
implemented in \holfour, learns to apply tactics extracted from existing proof 
scripts. It can perform induction on variables when an induction tactic has 
been defined for the particular inductive type. Yet, it is currently limited by 
its inability to synthesize terms as arguments of tactics.

\section{Conclusion and Future Work}
Our framework exhibits good performance on the two synthesis tasks exceeding 
the performance of state-of-the-art ATP on combinators solving $65\%$ of the 
test problems. Its success rate reaches $78.5\%$ on Diophantine equations.
Our proposed approach showcases how self-learning can solve a task by gathering 
examples from exploratory searches. Compared to supervised learning, this 
self-learning approach does not require the solutions of the problem to be 
known in advance.

In the future, we intend to test this reinforcement learning framework on
many more tasks and test the possibility of joint 
training~\cite{DBLP:journals/corr/abs-1902-04422}.
One domain to explore consists of other important tasks on 
higher-order terms such as beta-reduction or higher-order unification. 
Another interesting development is to apply the  
ideas of this paper to the \tactictoe framework. 
A direct application would 
give the tactical prover the ability to synthesize the terms appearing as 
arguments of tactic. 
In general, our framework is able to construct programs and therefore 
could be adapted to perform tactic synthesis.

\bibliographystyle{plain}
\bibliography{biblio}
\end{document}